\documentclass{article}
\usepackage{ijcai25}

\usepackage{times}
\usepackage{soul}
\usepackage{url}
\usepackage[hidelinks]{hyperref}
\usepackage[utf8]{inputenc}
\usepackage[small]{caption}
\usepackage{graphicx}
\usepackage{amsmath}
\usepackage{amsthm}
\usepackage{booktabs}
\usepackage{algorithm}
\usepackage{algorithmic}
\usepackage[switch]{lineno}
\usepackage{amssymb} 
\usepackage{multirow}
\usepackage[table,xcdraw]{xcolor}
\definecolor{bestcolor}{HTML}{F8CBAD}

\title{Leader and Follower: Interactive Motion Generation under Trajectory Constraints}

\author{
Runqi Wang$^{1,2}$
\and
Caoyuan Ma$^{1,2}$\and
Jian Zhao$^{3}$\and
Hanrui Xu$^{1,2}$\and
Dongfang Sun$^{1,2}$\and
Haoyang Chen$^{1,2}$\and
Lin Xiong$^{4}$\and
Zheng Wang$^{1,2}$\thanks{Corresponding author1}
\and
Xuelong Li$^{3}$\thanks{Corresponding author2}
\\
\affiliations
$^1$National Engineering Research Center for Multimedia Software, Institute of Artificial Intelligence\\
$^2$School of Computer Science, Wuhan University, China\\
$^3$The Institute of AI, China Telecom\\
$^4$Geely Auto Research\\
}
\definecolor{bestcolor}{HTML}{F8CBAD} 
\begin{document}

\twocolumn[{%
\renewcommand\twocolumn[1][]{#1}%
\vspace{-2mm}
\maketitle
\begin{center}
    \centering
    \vspace{-4mm}
    \includegraphics[width=\textwidth]{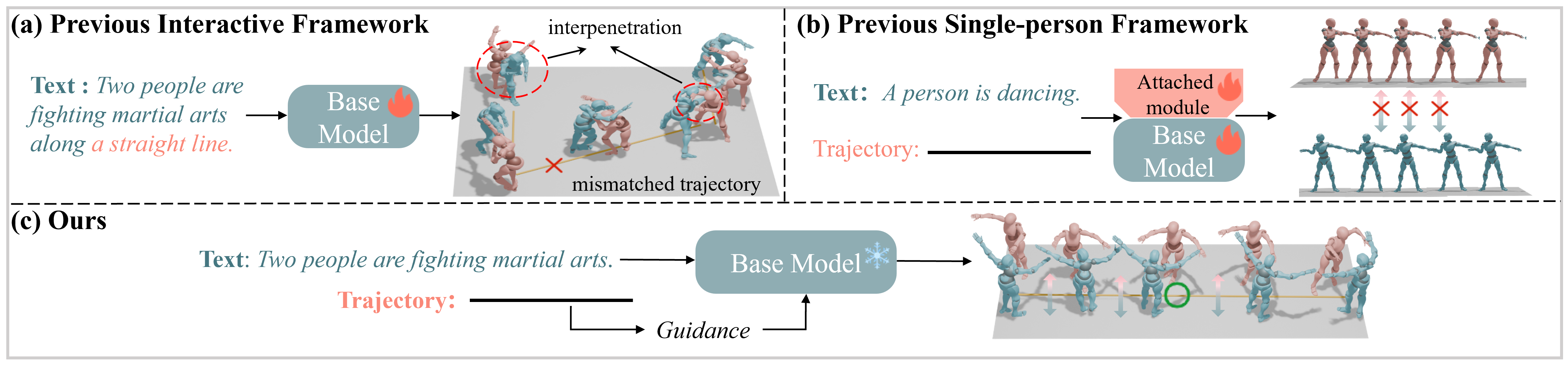}
     \vspace{-4mm}
      \captionof{figure}{\textbf{Comparison of Our Task with Previous Works.}   
     (a) Interaction methods based on textual input to describe trajectory result in trajectory deviations and interaction errors (as indicated by the red circle); 
     (b) Some methods for single-actor motion generation use 3D trajectories but require retraining and fail to account for inter-person interactions;
    (c) Our approach leverages precise 3D trajectory and textual input to guide interactive motion generation, achieving consistent trajectory generation without additional retraining.
      }
       \label{fig:teaser_1}
\end{center}%
}]

\begin{abstract}
With the rapid advancement of game and film production, generating interactive motion from texts has garnered significant attention due to its potential to revolutionize content creation processes.
In many practical applications, there is a need to impose strict constraints on the motion range or trajectory of virtual characters.
However, existing methods that rely solely on textual input face substantial challenges in accurately capturing the user's intent—particularly in specifying the desired trajectory.
As a result, the generated motions often lack plausibility and accuracy.
Moreover, existing trajectory-based methods for customized motion generation rely on retraining for single-actor scenarios, which limits flexibility and adaptability to different datasets, as well as interactivity in two-actor motions.
To generate interactive motion following specified trajectories,
this paper decouples complex motion into a Leader-Follower dynamic, inspired by role allocation in partner dancing.
Based on this framework, this paper explores the motion range refinement process in interactive motion generation and proposes a training-free approach, integrating
a Pace Controller and a Kinematic Synchronization Adapter.
The framework enhances the ability of existing models to generate motion that adheres to trajectory by controlling the leader's movement and correcting the follower's motion to align with the leader.
Experimental results show that the proposed approach, by better leveraging trajectory information, outperforms existing methods in both realism and accuracy.
\end{abstract}

\section{Introduction}


With advances in game and film production, generating interactive motion from textual descriptions has gained significant attention due to its potential to replace traditional motion capture and drive the transformation and enhancement of content creation workflows~\cite{guo2022generating,zhang2023generating,lin2023being}.
In practical applications such as fight choreography and animation production, characters often need to perform actions along specific trajectories or within designated areas. Therefore, generating motions that not only satisfy the textual pose requirements but also adhere to the specified trajectory has become an important research direction.




\begin{figure*}[ht]
    \centering
    \includegraphics[width=0.98\textwidth]{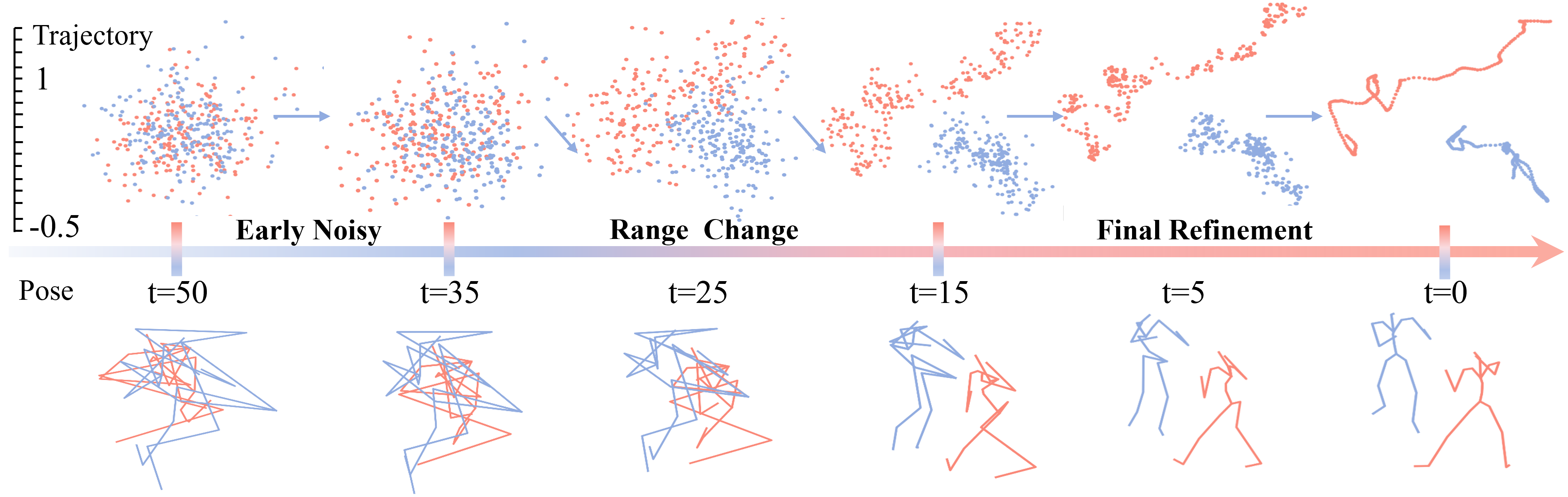}
    \caption{\textbf{Motion Range Refinement Process.} We visualize the trajectories and human poses at different time steps during the denoising process.
    The process of interactive motion generation is divided into three stages: early diffusion characterized by high noise and overlapping trajectories, mid-stage stabilization of movement direction and range, and final-stage refinement of motion details with stable motion range.}
    \label{fig:teaser_2}
\end{figure*}

Currently, methods for generating interactive motion conditioned on both text and trajectory can be broadly categorized into two types.
The first category directly incorporates trajectory descriptions into the text conditions (see Fig.\ref{fig:teaser_1}(a)). However, the text-based descriptions of trajectories are imprecise, and in multi-agent scenarios, issues such as interpenetration arise.
The second category consists of trajectory constraint methods for single-actor motion~\cite{wan2023tlcontrol,xie2023omnicontrol,karunratanakul2023guided,Song_2024_CVPR}, which typically require retraining based on different datasets or tasks (see Fig.\ref{fig:teaser_1}(b)). 
This design results in limited flexibility, increased computational costs, and a failure to adequately account for the interdependencies between roles in interactive movements.
In contrast, this study aims to guide motion diffusion models in generating interactive motions that satisfy the input action text and trajectory conditions, without the need for retraining (see Fig.\ref{fig:teaser_1}(c)).
To simplify this complex interactive task, we are inspired by the role allocation in two-person interactive motions and decoupled the interaction process, introducing the Leader-Follow paradigm. Specifically, we aim to first correct the trajectory of the leader, akin to the role of a lead dancer, and then adjust the behavior of the follower to coordinate with the leader's motion.

To guide the leader and refine the follower more accurately and effectively,  we investigated the underlying generation patterns and identified the Motion Range Refinement Process, as illustrated in Figure~\ref{fig:teaser_2}.
During the reverse denoising process, each stage focuses on different levels of information, reflecting a gradual transition from highly disordered noise to motion region formation and, finally, to detail generation.
We find that the mid-stage of diffusion is critical for trajectory adjustment, where the movement direction becomes defined, the motion range stabilizes, and the generated results gain semantic coherence, with the basic trajectory shape forming.
This phenomenon provides valuable insight.
We recognize that during the mid-stage of motion formation, timely trajectory adjustments help maintain consistency, mitigate early-stage noise disruptions, and prevent unnatural results from excessive adjustments as the motion range stabilizes.
Therefore, we consider the mid-stage to be critical for subsequent module adjustments.

Based on the above analysis, we choose to perform trajectory correction in the critical mid-stages of motion diffusion and realize our Lead-Follow Paradigm.
To ensure proper guidance of the leader within the pair, we introduce a unidirectional Pace Controller.
The motion diffusion process is bidirectional, meaning the trajectory is influenced by information from other joints during the denoising process.
Our unidirectional Pace Controller ensures that, at the mid-stage denoising step, the trajectory remains unaffected by information from other joints at both the beginning and end of each diffusion step, thereby guaranteeing the generated motion strictly adheres to the target trajectory distribution.
To ensure the coherence and consistency of the two-agent interactive motion, we design a Kinematic Synchronization Adapter that controls the Follower to align with the Leader.
This adapter performs motion domain detection and interaction correction based on the distance relationships between the Leader and the Follower, precisely adjusting the Follower's interactive motion. The corrected motion ensures adherence to both realistic physical principles and behavioral logic.

Our design represents the first exploration of the Motion Range Refinement Process in interactive motion, with the aim of constructing a modular, training-free universal framework. Extensive experiments demonstrate the effectiveness and feasibility of this design in the generation of interactive motion.
We hope this work could bring fundamental insights into related fields.

The contributions of this paper are threefold:
\begin{itemize}
\item This study is the first to investigate the guiding role of trajectory in interactive motion generation tasks,
decoupling the complex two-actor interactive motions and introducing the proposed Leader-Follower paradigm.
\item  This study explores the Motion Range Refinement Process in interactive motion generation and introduces the Pace Controller and Kinematic Synchronization Adapter to leverage trajectory guidance for generating interactive motion.

\item Extensive experimental results demonstrate that, compared to state-of-the-art methods, the proposed approach generates more realistic motion that adheres to both textual and trajectory conditions.
\end{itemize}

\begin{figure*}[ht]
    \centering
    \includegraphics[width=0.97\textwidth]{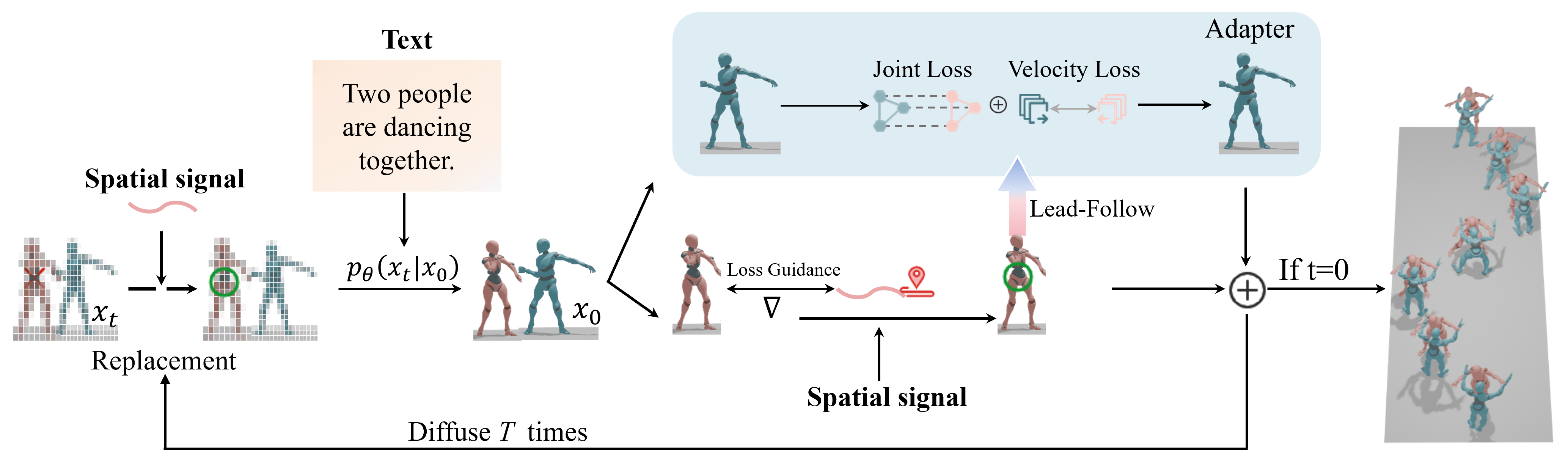}
      \vspace{-2mm}
    \caption{
    \textbf{Motion Generation Pipeline with Text and Trajectory input.}
    Inspired by partner dance leadership, we first use a Controller to define the leader's trajectory, and then employ an Adapter to guide the follower's motion to align with the leader.
    }
    \label{fig:mehtod}
\end{figure*}

\section{Related Work}
\subsection{Interactive Motion Generation}
Recently, various methodologies~\cite{liang2024intergen,zhou2023unified,wang2023intercontrol,ghosh2024remos,xu2024inter,Lee2024CVPR,shan2024towards,Li_2024_CVPR,jang2024geometry} have been developed for Interactive Motion Generation with humans.
However, existing methods primarily depend on textual prompts to guide motion generation, which limits their ability to control complex interactive movements, especially in terms of trajectory control. 
Our approach leverages key trajectory constraints to enable precise control over interactive motion generation, enhancing adaptability to complex scenarios and aligning generated motions with diverse space demands in real-world situations.
\subsection{Trajectory Guided Motion Generation}
In the field of single-person motion generation, prior studies have explored the integration of trajectories into the human motion generation process.
GMD~\cite{karunratanakul2023guided} proposed a two-stage framework that first generates trajectories satisfying keyframe constraints and then synthesizes complete motions.
OmniControl~\cite{xie2023omnicontrol} demonstrated the capability to control multiple joints using a single model, but it may produce unnatural motions when spatial control signals for different joints conflict.
TLControl~\cite{wan2023tlcontrol} employs a VQ-VAE~\cite{van2017neural} and a Masked Trajectory Transformer to predict motion distributions based on user-specified partial trajectories and textual descriptions.
However, these approaches are confined to single-person motion generation and fail to account for the interactive dynamics of multi-person movements.
Moreover, they often require redesigning and retraining models for different tasks, architectures, or new datasets.
We pioneer the exploration of diffusion models in the context of interactive-person motion generation, introducing trajectories into this domain for the first time without necessitating any additional retraining of existing models.

\section{Method}
\subsection{Preliminary}
Diffusion-based probabilistic generative models (DPMs) are modeled as a Markov noising process, where the input \( x_t \) is progressively noised with varying noise levels \( t \). The noising process is defined cumulatively as
\(
 q(x_t | x_0) = \mathcal{N} \left( \sqrt{\alpha_t} x_0, (1 - \alpha_t) I \right)
\),
where \( x_0 \) is the clean input, \( \alpha_t = \prod_{s=1}^t (1 - \beta_s) \), and \( \beta_t \) is a noise scheduler.
The denoising model \( p_\theta(x_{t-1} | x_t) \) with parameters \( \theta \) learns to reverse the noising process by modeling the Gaussian posterior distribution \( q(x_{t-1} | x_t, x_0) \). 
After \( T \) successive denoising steps, diffusion models can map a prior distribution \( \mathcal{N}(0, I) \)to any target distribution \( p(x) \).

We model a two-person interaction \( x \) as a collection
of two single-person motion sequences \( x_h \), i.e., \( x = \{ x_a, x_b \} \), where \( x_h = \{ x_i \}^{L}_{i=1} \) is a fixed-framerate sequence of motion poses \( x_i \).
Multi-Human Motion Generation~\cite{liang2024intergen} is based on a fundamental assumption, \textbf{commutative property}, which means that two-person \(\{x_a, x_b\}\) and \(\{x_b, x_a\}\) are equivalent, the order of every single motion does not change the semantics of the interaction itself, the distribution of interaction data satisfies the following property:\(p(x_a, x_b) \equiv p(x_b, x_a) \).
Based on this principle, Multi-Human Motion Generation ~\cite{liang2024intergen} addresses the symmetry of human identities in interaction by utilizing cooperative transformer-based denoisers with shared weights.

The base model InterGen~\cite{liang2024intergen} is designed to ensure that interactive motion adheres to physical laws and its process characteristics (such as interactivity).
To ensure physical plausibility and consistency, geometric loss functions such as foot contact, joint velocity, and bone length loss are used, following the approach in MDM~\cite{tevet2023human}. Additionally, interactive losses, including masked joint distance map (DM) loss and relative orientation (RO) loss, are introduced to address spatial complexity in Multi-Human interactions~\cite{liang2024intergen}.


\subsection{Pace Controller}
\begin{algorithm}[tb]
    \caption{Pace Controller}
    \label{alg:TCR}
    \textbf{Input}: Initial motion $x_t = \{{x}^t_a,{x}^t_b \}$, ground truth trajectories $x_a^{\text{proj}}$, trajectory period $[T_1, T_2]$, Trajectory loss function $G_t(\cdot)$\\
    \textbf{Parameter}:Motion diffusion model $M_{\theta}$\\
    \textbf{Output}:Motion $x_0 = \{{x}^0_a,{x}^0_b \}$
    \begin{algorithmic}[1] 
         \FOR{$t = 1$ to $T$}
             \IF{$t \in [T_1, T_2]$}
                \STATE $x_t = \{{x}^t_a,{x}^t_b \} \leftarrow x_a^{\text{proj}}(t)$
            \ENDIF
            \STATE $x_{0} \leftarrow M_{\theta}(x_{t})$
            \STATE $x_{0} \leftarrow \nabla_{x_{0}} G_t(x_a^{\text{proj}},x_a^{\text{0proj}})$
            \STATE $ \mu, \Sigma \leftarrow \mu(x_0, x_t), \Sigma_t$
            \STATE $x_{t-1} \sim \mathcal{N} \left( \mu , \Sigma \right)$
        \ENDFOR
        \STATE \textbf{return} Optimized motion $x_0 = \{{x}^0_a,{x}^0_b \}$.
    \end{algorithmic}
\end{algorithm}
The Pace Controller aims to guide the leader's trajectory to align with the input conditions. However, the bidirectional data diffusion contradicts our goal of unidirectional guidance.
Specifically, inspired by nonequilibrium thermodynamics, the diffusion model establishes a Markov chain between the target data distribution and a Gaussian distribution.
In this process, data interactions between the root joint and other joints influence each other (i.e., the root joint data is also modified by other joints), gradually transitioning from the Gaussian distribution to the target distribution.
Therefore, the core challenge of Pace Controller lies in effectively ensuring the accuracy of leader's root joint trajectories amidst such interactions, protecting them from interference by other joint components, and leveraging trajectory information to guide and refine the generation of leader motion effectively.

Based on the above analysis, we propose a unidirectional diffusion-guided strategy to ensure the leader's trajectory consistency.
During the trajectory formation phase, we employ a strategic intervention to directly replace the relevant portion of \( x_t \) at time step $t$ with the real trajectory segment from the input condition  $x_a^{\text{proj}}$. This replacement ensures that the trajectory segment associated with the leader is maintained with absolute accuracy at the start of the denoising process.

\begin{equation}
{x}_t = \left(  x_t \leftarrow x_a^{\text{proj}}\right) \cdot \mathbb{I}_{[T_1, T_2]}(t) + x_t \cdot \mathbb{I}_{[T_1, T_2]^c}(t),
\end{equation}

where \( \mathbb{I}_{[T_1, T_2]}(t) \) equals 1 if \( t \in [T_1, T_2] \) and 0 otherwise, while \( \mathbb{I}_{[T_1, T_2]^c}(t) \) equals 1 if \( t \notin [T_1, T_2] \) and 0 otherwise.

Moreover, throughout the reverse denoising process, for each predicted clean state $x_0$ at time step $t$, we extract the trajectory portion and compute the per-frame MSE loss with respect to the given trajectory condition, thereby optimizing the current state prediction. 
After optimization, the updated state $x_0$ is further involved in the diffusion process, leading to an improvement in the accuracy of the final generated trajectory.
\begin{equation}
x_0 = \mathcal{G}_\text{opt} \left( x_0; ||x_0 - x_a^{\text{proj}}(t)|| ^2 \right),
\end{equation}

\(\mathcal{G}_\text{opt}\)  represents the optimization function, applied to minimize \(\text{MSE}\) loss and update the current state \(x_0\).


\subsection{Kinematic Synchronization Adapter}

Forcibly modifying the trajectory of the leader during an interaction may adversely affect the interactivity and compatibility of both participants.
To address this issue, we have designed the Kinematic Synchronization Adapter, which leverages the interactive relationship between two participants to ensure the motion of the follower remains consistent and reasonable.
Specifically, we first develop a conflict detection module that utilizes the SMPL model~\cite{loper2023smpl} to delineate the interaction domain between two bodies.
\begin{equation}
\mathcal{C}_t = \mathbb{I}_{\left|\mathcal{D}_a^t \cap \mathcal{D}_b^t\right| > 0}
\end{equation}

where \(\mathcal{D}_a^t = \mathcal{M}(\mathbf{x}_a^t) \) and \(\mathcal{D}_b^t = \mathcal{M}(\mathbf{x}_b^t)\) represent the interaction domains of Leader \({x}^t_a\) and Follower \({x}^t_b\), at time step \(t\), obtained through the SMPL model mapping.
Upon detecting an overlap within this domain, the position state of the follower is adjusted.
\begin{equation}
x^t_b = \mathbb{I}_{\mathcal{C}_t = \emptyset} \left( x^t_b \right) + \mathbb{I}_{\mathcal{C}_t \neq \emptyset} \left( \mathcal{F}_\text{opt} \left( x^t_b; \mathcal{S}(x^t_a,x^t_b) \right) \right)
\end{equation}

$\mathbb{I}_{\mathcal{C}_t \neq \emptyset}$ and $\mathbb{I}_{\mathcal{C}_t = \emptyset}$ are indicator functions.
The optimal adjustment function $\mathcal{F}_\text{opt}$ modifies the trajectory of the follower based on the Leader-Follower interactive relationship $\mathcal{S}(\mathbf{x}_a^t, \mathbf{x}_b^t)$ between the participants.
We still opt for 2-3 denoising steps during the trajectory formation phase, applying reverse motion to the detected colliding individuals first and then imposing a joint distance loss function to constrain the relative position between the two individuals.
The distance loss function may lead to highly similar human movements at the end frame of the motion. Therefore, we introduce a discrepancy velocity loss function as a penalty term.
\begin{equation}
\mathcal{S}(x^t_a,x^t_b) = \mathcal{L}_{\text{joint}}+\mathcal{L}_{\text{velocity}}
\end{equation}
\begin{equation}
\mathcal{L}_{\text{joint}}
=\sum_{j} \left\|  \max(0, \delta - \| \mathbf{p}_a(t)(j) - \mathbf{p}_b(t)(j) \|_2)^2 \right\|^2
\end{equation}
\begin{small}
\begin{equation}
\mathcal{L}_{\text{velocity}} = \sum_{t=1}^{n-1} \sum_{j=1}^{22} \frac{\|\mathbf{v}_a(t,j)\|_2 \|\mathbf{v}_b(t,j)\|_2}{\|\mathbf{v}_a(t,j)\|_2 \|\mathbf{v}_b(t,j)\|_2} (\mathbf{v}_a(t,j) \cdot \mathbf{v}_b(t,j))
\end{equation}
\end{small}

The function \( \mathcal{S}(x_a^t, x_b^t) \) represents the combined loss function, consisting of the joint distance loss \( \mathcal{L}_{\text{joint}} \) and the velocity loss \( \mathcal{L}_{\text{velocity}} \).
The joint distance loss is used to constrain the relative positions of the Leader and Follower at each time step, ensuring that the joints of the two participants maintain an appropriate distance throughout the interaction. 
The velocity loss applies a similarity penalty to prevent the homogenization of the two participants in the final frames, thus preserving the dynamic nature of the interaction.

\section{Experiment}

\begin{small}
\begin{table*}[t]
\centering
\resizebox{\textwidth}{!}{
\begin{tabular}{lccccccc}
\toprule
\multirow{2}{*}{\small \textbf{Methods}} & \multicolumn{3}{c}{\small \textbf{R-Precision $\uparrow$}} & \multirow{2}{*}{\small \textbf{FID ↓}} &\multirow{2}{*}{\small \textbf{MM Dist ↓}} & \multirow{2}{*}{\small \textbf{Diversity →}} & \multirow{2}{*}{\small \textbf{MModality ↑}} \\
\cmidrule(lr){2-4}
& \small Top 1 & \small Top 2 & \small Top 3 & & & & \\
\midrule
\small TEMOS~(\citeauthor{petrovich2022temos})  & 0.224\textsuperscript{$\pm$.010} & 0.316\textsuperscript{$\pm$.013} & 0.450\textsuperscript{$\pm$.018} & 17.375\textsuperscript{$\pm$.043} & 5.342\textsuperscript{$\pm$.015} & 6.939\textsuperscript{$\pm$.071} & 0.535\textsuperscript{$\pm$.014} \\
\small T2M~(\citeauthor{guo2022generating})  & 0.238\textsuperscript{$\pm$.012} & 0.325\textsuperscript{$\pm$0.11} & 0.464\textsuperscript{$\pm$0.14} & 13.769\textsuperscript{$\pm$0.72} & 4.731\textsuperscript{$\pm$0.13} & 7.046\textsuperscript{$\pm$0.22} & 1.387\textsuperscript{$\pm$0.76} \\
\small MDM~(\citeauthor{tevet2023human})  & 0.153\textsuperscript{$\pm$.012} & 0.260\textsuperscript{$\pm$.009} & 0.339\textsuperscript{$\pm$.012} & 9.167\textsuperscript{$\pm$.056} & 6.125\textsuperscript{$\pm$.018} & 7.602\textsuperscript{$\pm$.045} & 2.355\textsuperscript{$\pm$.080} \\
\small ComMDM*~(\citeauthor{shafir2023human}) & 0.067\textsuperscript{$\pm$.013} & 0.125\textsuperscript{$\pm$.018} & 0.184\textsuperscript{$\pm$.015} & 38.643\textsuperscript{$\pm$.098} & 13.211\textsuperscript{$\pm$.013} & 3.520\textsuperscript{$\pm$.058} & 0.217\textsuperscript{$\pm$.018} \\
\small ComMDM~(\citeauthor{shafir2023human})  & 0.223\textsuperscript{$\pm$.013} & 0.334\textsuperscript{$\pm$.011} & 0.466\textsuperscript{$\pm$.014} & 7.069\textsuperscript{$\pm$.054} & 5.212\textsuperscript{$\pm$.012} & 7.244\textsuperscript{$\pm$.038} & 1.822\textsuperscript{$\pm$.052} \\
RIG~(\citeauthor{tanaka2023role})  & 0.285\textsuperscript{$\pm$.012} & 0.409\textsuperscript{$\pm$.014} & 0.521\textsuperscript{$\pm$.012} & 6.775\textsuperscript{$\pm$.069} & 4.876\textsuperscript{$\pm$.018} & 7.311\textsuperscript{$\pm$.043} & 2.096\textsuperscript{$\pm$.065} \\
\small InterGen~(\citeauthor{liang2024intergen})  & 0.371\textsuperscript{$\pm$.010} & 0.515\textsuperscript{$\pm$.012} & 0.624\textsuperscript{$\pm$.012} & 5.918\textsuperscript{$\pm$.079} & 4.108\textsuperscript{$\pm$.014} & 7.387\textsuperscript{$\pm$.029} & 2.141\textsuperscript{$\pm$.063} \\
\midrule
 \cellcolor{bestcolor}\textbf{Ours}& \cellcolor{bestcolor} \textbf{0.522\textsuperscript{$\pm$.006}} & 
 \cellcolor{bestcolor}\textbf{0.658\textsuperscript{$\pm$.004}}&
 \cellcolor{bestcolor}\textbf{0.737\textsuperscript{$\pm$.005}}&
 \cellcolor{bestcolor}\textbf{5.352\textsuperscript{$\pm$.076}}&
 \cellcolor{bestcolor}\textbf{3.778\textsuperscript{$\pm$.001}}&
 \cellcolor{bestcolor}\textbf{7.931\textsuperscript{$\pm$.036}}&
 \cellcolor{bestcolor}1.174\textsuperscript{$\pm$.026}
\\
\bottomrule
\end{tabular}
}
\vspace{-2mm}
\caption{\textbf{Quantitative Comparisons of Various Methods.} Our method outperforms existing approaches in R-Precision, FID, MM Distance, and Diversity, demonstrating superior motion generation quality, greater diversity, and better alignment with textual conditions.}
\label{tab:ComparisonWithSota}
\end{table*}
\end{small}

\begin{figure*}[ht!]
    \centering
    \includegraphics[width=\textwidth]{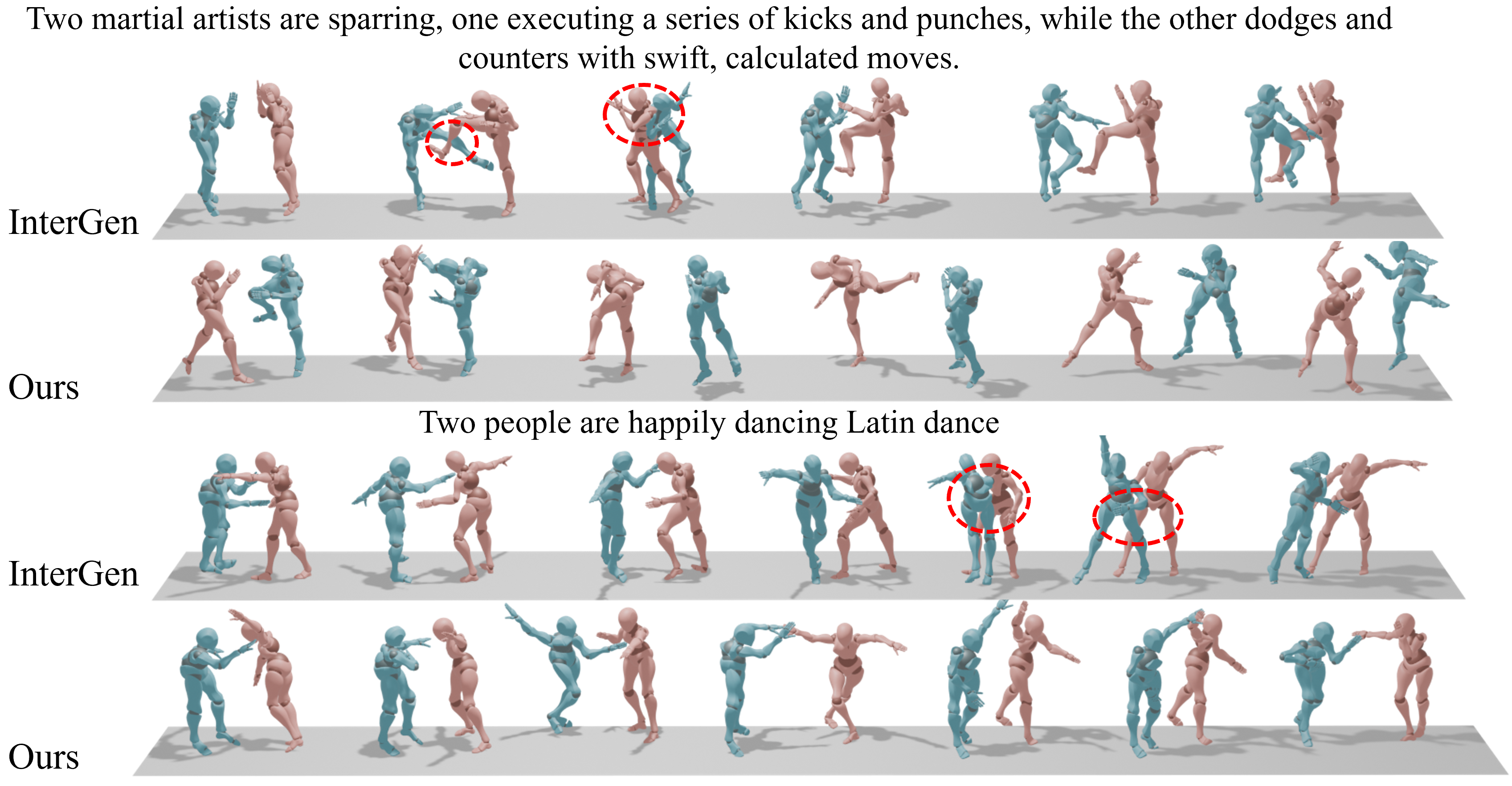}
      \vspace{-3mm}
    \caption{\textbf{Visual Comparison with Other Methods.} In complex scenarios requiring close contact and interaction, baseline models often produce unnatural interpenetration (as indicated by the~\textcolor{red}{red} circles). Our approach controls the leader's trajectory and guides the follower's actions to align with the leader, thereby effectively addressing these issues.
    }
    \label{fig:vis_com}
\end{figure*}

\begin{figure*}[ht!]
    \centering
    \includegraphics[width=\textwidth]{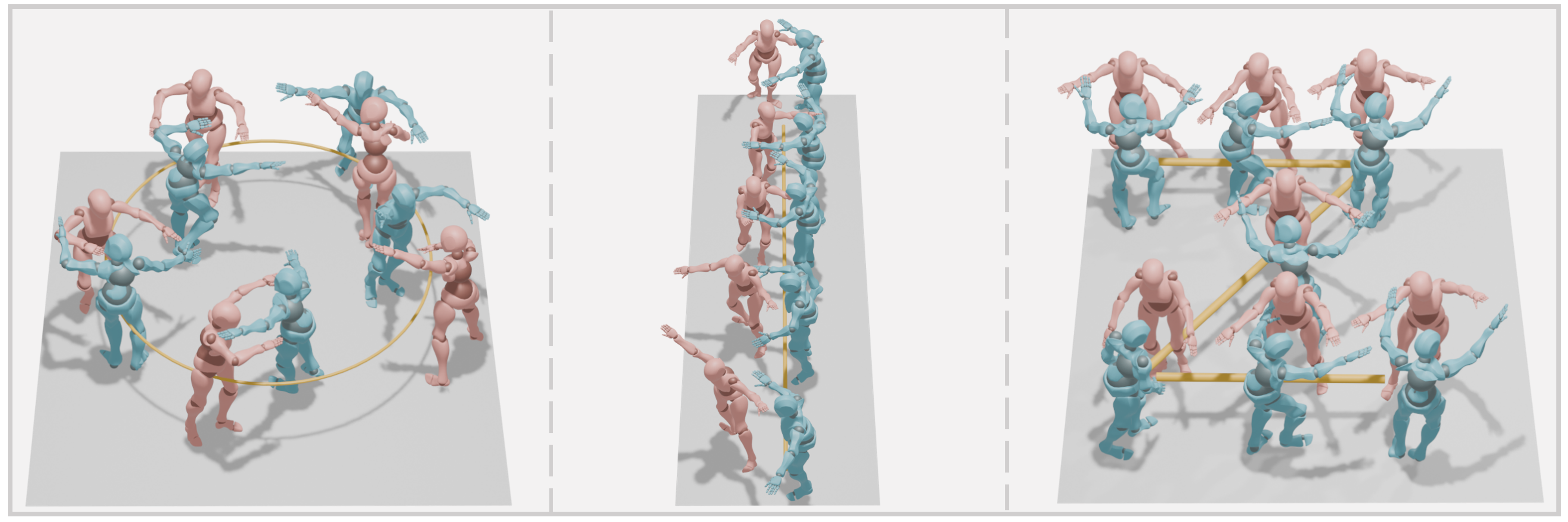}
      \vspace{-5mm}
    \caption{\textbf{Demonstration of Trajectory Guidance Effect.} For the same input text, ``Two people are dancing together,'' we provide different trajectory conditions. All generated sequences align with both the trajectory and textual features, resulting in realistic and natural motions.
 }
    \label{fig:effect_of_trajectory_guidance}
\end{figure*}

\textbf{Evaluation dataset.} We utilize the InterHuman~\cite{liang2024intergen} as our primary dataset. This dataset comprises 7,779 motions spanning various categories of human motions, annotated with 23,337 unique descriptions containing 5,656 distinct words, and encompassing a total duration of 6.56 hours.

\noindent\textbf{Evaluation metrics.} We adopt the evaluation metrics consistent with InterGen~\cite{liang2024intergen}, which are listed as follows:
\begin{enumerate}
    \item \textbf{R-Precision.} To measure the text-motion consistency, we rank the Euclidean distances between the motion and text embeddings. Top-1, Top-2, and Top-3 accuracy of motion-to-text retrieval are reported.
    \item \textbf{Frechet Inception Distance (FID).} To measure the similarity between synthesized and real interactive motions, we calculate the latent embedding distribution distance between the generated and real interactive motions using FID on the extracted motion features.
    \item \textbf{Multimodal Distance (MM Dist).} To measure the similarity between each text and the corresponding motion, the average Euclidean distance between each text embedding and the generated motion embedding from this text is reported.
    \item \textbf{Diversity.} We randomly sample 300 pairs of motions and calculate the average Euclidean distances of the pairs in latent space to measure motion diversity in the generated motion dataset.
    \item \textbf{Multimodality (MModality).} Similar to Diversity, we sample 20 motions within one text prompt to form 10 pairs, and measure the average latent Euclidean distances of the pairs. The average over all the text descriptions is reported.
\end{enumerate}

\subsection{Implementation Details.}
We implement InterGen~\cite{liang2024intergen} as the baseline model, which consists of \( N = 8 \) blocks, each with a latent dimension of 1024.
Each attention layer within the model is composed of 8 heads.
During training, the number of diffusion timesteps is set to 1,000.
For sampling, we apply the DDIM~\cite{song2020denoising} strategy with 50 timesteps and set \( \eta = 0 \).
Additionally, a frozen CLIP-ViTL/14 model is employed as the text encoder.
All experiments are conducted on an Nvidia RTX 3090 GPU.

\subsection{Comparison with State-of-the-art Methods}
Table~\ref{tab:ComparisonWithSota} presents a comparison between our method and other state-of-the-art approaches across multiple dimensions.
In terms of R-Precision, our method significantly outperforms existing methods across all Top-K evaluations, indicating that the trajectory-guided signal provides clearer spatiotemporal alignment information, which effectively improves the matching between the generated motion and the target reference.
Regarding the FID metric, our method further reduces to 5.352 compared to InterGen~\cite{liang2024intergen}, highlighting the crucial role of the trajectory signal in guiding interactive motion generation. This signal enhances both the consistency and realism of the generated motion.
In terms of diversity, our method achieves high diversity scores, suggesting that although the trajectory imposes some constraints on the generated samples, it remains a low-dimensional signal and does not limit the diversity of the generated motions.
However, the lower MM modality score of 1.174 indicates that while trajectory guidance enhances the spatiotemporal consistency of the generated samples, it also somewhat suppresses the divergence of the multimodal distribution in the generated results.

\subsection{Qualitative Results}

\noindent \textbf{Analysis of Visual Comparison with Other Methods.}
Figure~\ref{fig:vis_com} illustrates the visual results of our method compared to other approaches.
In complex scenarios requiring close contact and interaction, such as martial arts combat and Latin dance, the baseline model generates unnatural overlaps, intersections, or penetrations (highlighted by red circles), compromising motion authenticity and fluidity. In the martial arts scene, overlapping actions between characters violate physical and behavioral logic, while in the Latin dance scene, penetration and distortion disrupt the dance’s fluidity and aesthetic. By incorporating trajectory range and distance constraints, our method effectively resolves these issues, ensuring spatiotemporal consistency in the generated motions. The resulting actions are not only more natural and realistic but also adhere to physical laws and behavioral logic, exhibiting enhanced interactivity and coherence.

\noindent \textbf{Analysis of Trajectory Guidance.}
Figure~\ref{fig:effect_of_trajectory_guidance} presents the motion sequences generated under the same text ({\itshape Two people are dancing together}) with different trajectory signals for guidance. The experimental results indicate that the proposed model effectively leverages various trajectory signals to generate motion sequences that align with both specific textual and trajectory features. By incorporating diverse trajectory conditions, the model not only generates richer motion sequences but also ensures that the introduction of trajectory signals does not cause distortion or unreasonable outcomes. Even under trajectory constraints, the generated sequences maintain high-quality fluidity and consistency, demonstrating an excellent balance between diversity and consistency, surpassing the capabilities of existing methods.

\subsection{Ablation Studies}
\begin{figure}[t!]
    \centering
    \includegraphics[width=0.48\textwidth]{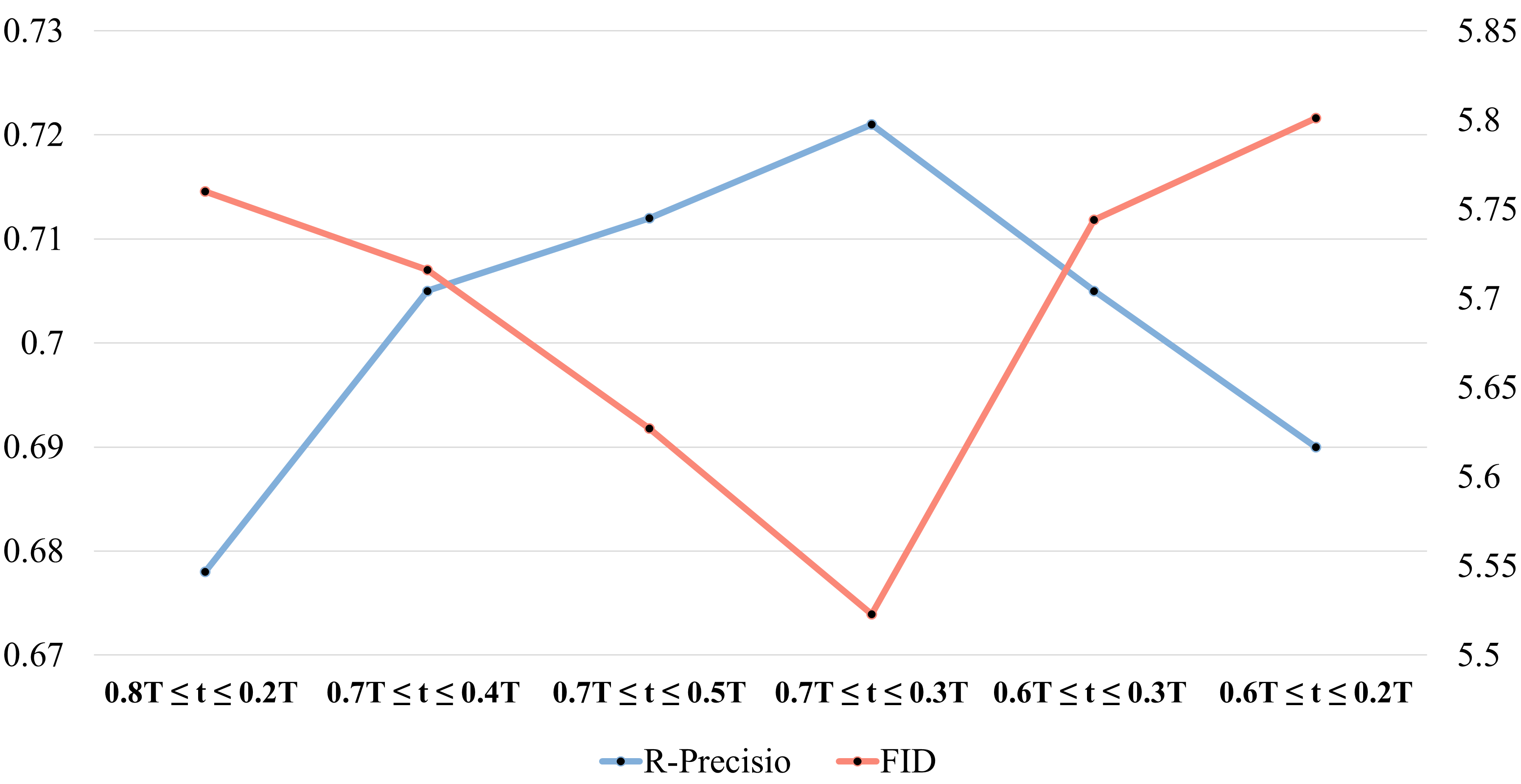}
      \vspace{-4mm}
    \caption{\textbf{Impact of different trajectory-guiding period on the Pace Controller.} Results show that covering the mid-stage of the Motion Range Refinement Process ($0.7T \leq t \leq 0.3T$) achieves the best motion authenticity and accuracy.
 }
    \label{fig:trjc_period_effect}
\end{figure}
\textbf{Analysis of the Trajectory Guidance Period.}
Figure~\ref{fig:trjc_period_effect} illustrates the impact of different trajectory guidance periods on the final generated results. We selected several time intervals during the denoising process as trajectory guidance and reported R-Precision and FID metrics. The results indicate that when the trajectory guidance period is set to the trajectory formation phase ($0.7T \leq t \leq 0.3T$), the generated results exhibit optimal performance in both semantic consistency (R-Precision) and action realism (FID). In contrast, when trajectory guidance is applied during the early stage of the denoising process ($0.8T \leq t \leq 0.2T$), the accumulated redundant noise during this stage leads to chaotic generation, significantly reducing semantic consistency. Additionally, when the trajectory guidance period is too short ($0.6T \leq t \leq 0.4T$), it is insufficient to provide the necessary semantic guidance, resulting in a decrease in the alignment between the generated actions and the input conditions.

\noindent \textbf{Analysis of Trajectory Guidance of Different Individuals.}
Table~\ref{tab:Trajectory Guidance for Different Individuals} shows the results of motion generation for the trajectories of different individuals as input conditions.
Due to the equivalence of two-actor motion, using the trajectory of either individual as the input condition results in minimal differences.
In this case, the FID and R-Precision yield consistent results, achieving both realistic motion generation and a high degree of textual alignment.
However, when both individuals' trajectories are used as input conditions simultaneously, FID and R-Precision worsened. This is because the constraints imposed on both actors' conditions are excessively rigid, potentially causing a misalignment with the textual description, leading to unnatural generation outcomes such as model clipping. Therefore, further refinement of the constraint conditions is necessary to ensure the coherence and naturalness of the interactive motion.
\begin{small}
\begin{table}[t]
    \centering
    \renewcommand{\arraystretch}{1.3} 
    \begin{tabular}{ccccc}
        \hline
        \small \textbf{$Person_1$} & \small \textbf{$Person_2$} & \small \textbf{FID$\downarrow$} & \small \textbf{R-Precision$\uparrow$} & \small \textbf{Diversity$\uparrow$} \\ \hline
        {\large $\circ$} & {\large $\circ$} & 5.918 & 0.624 & 7.387\\ 
        {$\bullet$} & {\large $\circ$} & 5.476 & 0.719 & 7.625 \\ 
        {\large $\circ$} & {$\bullet$} & 5.492 & 0.716 & 7.924 \\ 
        {$\bullet$} & {$\bullet$} &5.821& 0.684 & 7.524\\ \hline
    \end{tabular}
    \vspace{-3mm}
    \caption{\textbf{Trajectory Guidance for Different Individuals.}
    The performance is similar when using the leader's or follower's trajectory as input. However, it decreases when both individuals' trajectories are used simultaneously.}
    \label{tab:Trajectory Guidance for Different Individuals}
\end{table}
\end{small}

\begin{figure}[t]
    \centering
    \vspace{-3mm}
    \includegraphics[width=0.48\textwidth]{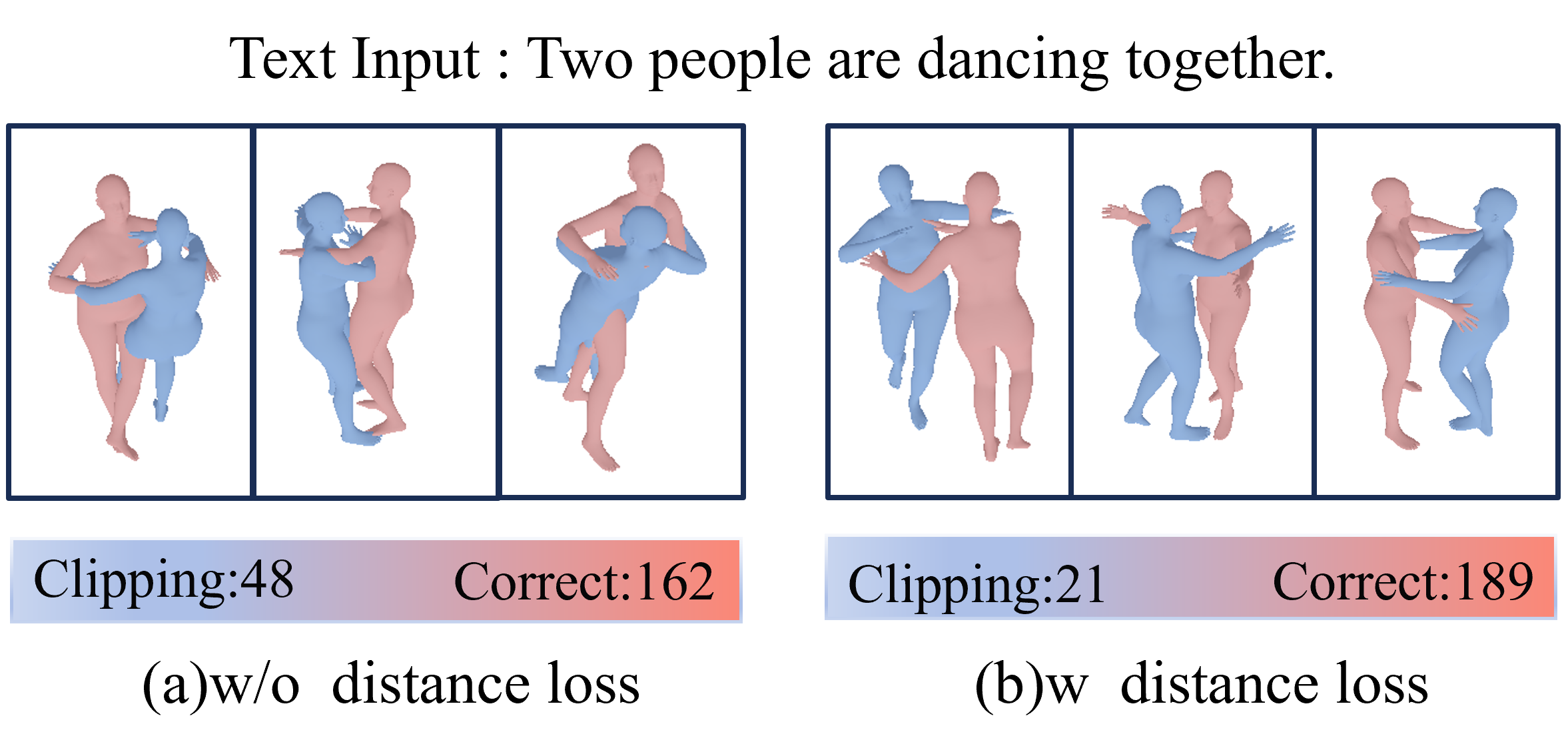}
      \vspace{-6mm}
    \caption{\textbf{Comparison experiment of Joint Loss.} 
    Joint Loss effectively adjusts the follower's position, reducing model clipping.}
    \label{fig:skin_loss}
\end{figure}

\noindent \textbf{Analysis of Joint Loss.}
Figure~\ref{fig:skin_loss} illustrates the corrective effect of Joint Loss within our Kinematic Synchronization Adapter.
We select a textual description with close contact between individuals.
The introduction of Joint Loss effectively constrains the spatial relationship between the two, improving the plausibility of relative joint positions and preventing potential collisions in intermediate frames during the motion correction by the Pace Controller. To quantitatively evaluate this effect, we employed a collision detection module to analyze the number of penetration frames across 210 frames of motion before and after correction. Our correction reduce the number of penetration frames by nearly half.

\begin{figure}[ht]
    \centering
    \vspace{-4mm}
    \includegraphics[width=0.48\textwidth]{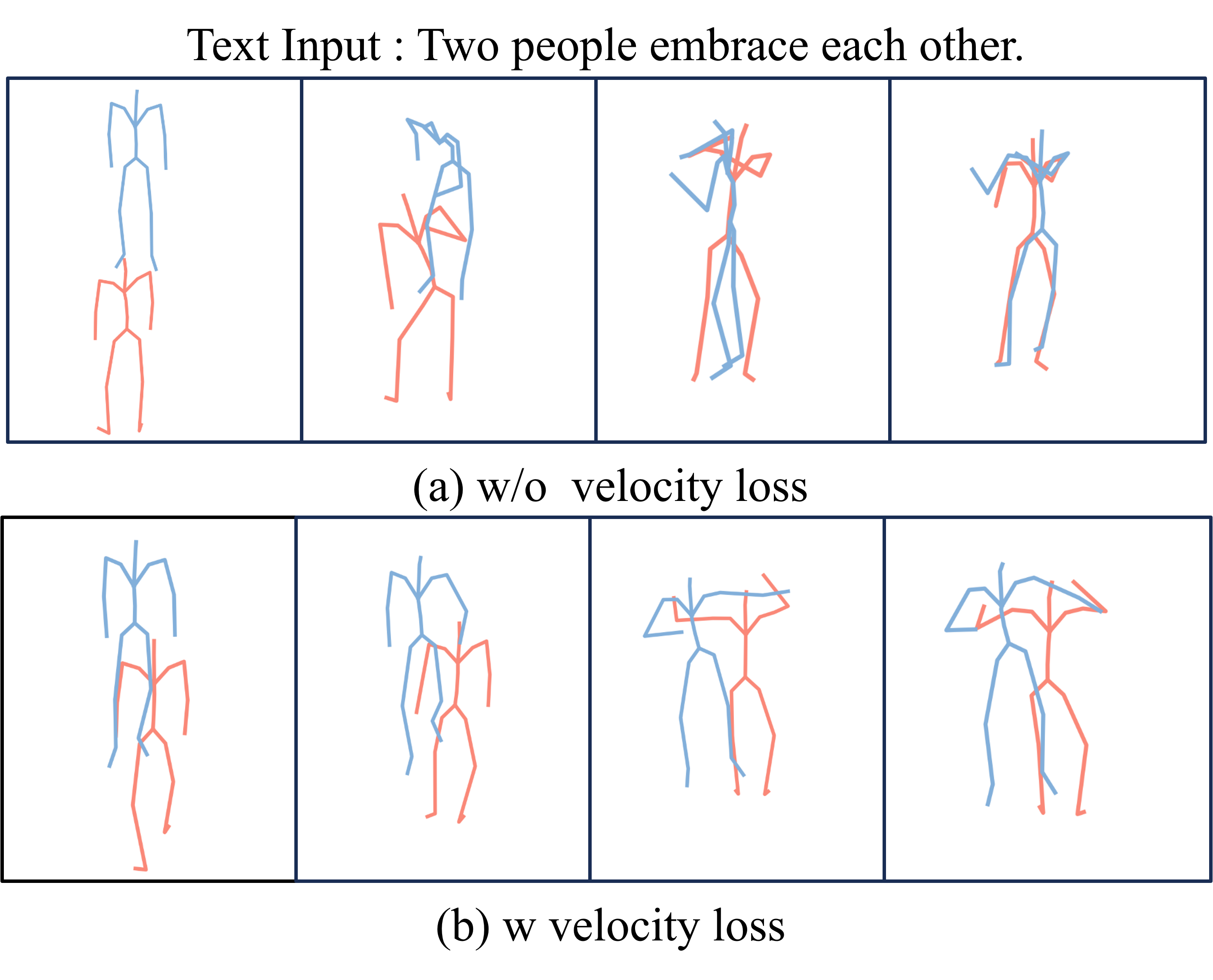}
      \vspace{-8mm}
    \caption{\textbf{Comparison experiment of velocity loss .}
    Velocity loss effectively prevents the follower's motion from becoming too similar to the leader's in the final few frames.}
    \label{fig:vel_loss_com1}
\end{figure}

\noindent \textbf{Analysis of Velocity Loss.}
Figure~\ref{fig:vel_loss_com1} (a) illustrates the effect of the Velocity Loss correction within the Kinematic Synchronization Adapter. When only joint loss is applied, the experiment reveals an issue in the final frames: the corrected character undergoes an abrupt reversal, and the motions of both characters gradually converge.
Upon introducing Velocity Loss, as shown in Figure~\ref{fig:vel_loss_com1} (b), the diversity penalty effectively prevents the homogenization of the last frames, ensuring that the movements remain natural and coherent.

\noindent \textbf{Analysis of Inference Time.}
Table~\ref{tab:inference_time} presents results regarding inference time. We randomly select the same 20 prompts, generate 210 frames for each movement, and perform the computation using an Nvidia RTX 3090 GPU, reporting the average inference time.
The base model employs the DDIM sampling method, which facilitates rapid generation. The introduction of Pace Controller and Kinematic Synchronization Adapter does not significantly increase the computational overhead during inference; the inference time only increases marginally, by approximately four seconds.
\begin{small}
\begin{table}[ht]
    \centering
    \renewcommand{\arraystretch}{1.3} 
    \begin{tabular}{ccccc}
        \hline
        \small \textbf{Model} & \small \textbf{Frame} & \small \textbf{Controller} & \small \textbf{Adapter} & \small \textbf{Inference Time} \\ \hline
        {$\bullet$} & 210 & {\large $\circ$} & {\large $\circ$} & 12.242\\ 
        {$\bullet$} & 210 & {$\bullet$} & {\large $\circ$} & 12.916 \\ 
        {$\bullet$} & 210 & {\large $\circ$} & {$\bullet$} & 14.281 \\ 
        {$\bullet$} & 210 & {$\bullet$} & {$\bullet$} & 16.169 \\ \hline
    \end{tabular}
    \vspace{-3mm}
    \caption{\textbf{Inference time comparison.}
Our design does not significantly increase the inference time cost.}
    \label{tab:inference_time}
\end{table}
\end{small}

\section{Conclusion}
This paper investigates the problem of generating multi-person interactive motions based on textual descriptions and 3D trajectory conditions.
Inspired by the role allocation in partner dances, the study introduces a Lead-Follow Paradigm and provides an in-depth analysis of the motion range optimization process in motion diffusion models.
A novel framework is proposed that operates without the need for retraining, incorporating both the Pace Controller and the Kinematic Synchronization Adapter.
Extensive experiments demonstrate that the proposed framework significantly outperforms existing methods and shows substantial improvements in the plausibility and accuracy of the generated motions.

\clearpage
\appendix

\bibliographystyle{named}
\bibliography{ijcai25}

\end{document}